# Network Threat Detection: Addressing Class Imbalanced Data with Deep Forest

Jiaqi Chen
Independent Researcher
Chicago IL, USA
ronanchen@google.com

Rongbin Ye
Independent Researcher
Chicago IL, USA
Rongbin.ye.94@gmail.com

*Abstract*—With the rapid expansion of Internet of Things (IoT) networks, detecting malicious traffic in real-time has become a critical cybersecurity challenge. This research addresses the detection challenges by presenting a comprehensive empirical analysis of machine learning techniques for malware detection using the IoT-23 dataset provided by the Stratosphere Laboratory. We address the significant class imbalance within the dataset through three resampling strategies. We implement and compare a few machine learning techniques. Our findings demonstrate that the combination of appropriate imbalance treatment techniques with ensemble methods, particularly gcForest, achieves better detection performance compared to traditional approaches. This work contributes significantly to the development of more intelligent and efficient automated threat detection systems for IoT environments, helping to secure critical infrastructure against sophisticated cyber attacks while optimizing computational resource usage.

## I. INTRODUCTION

Network attacks and cybersecurity have become increasingly prevalent and sophisticated in recent years, posing significant threats to individuals, organizations, and critical infrastructure. The proliferation of Internet of Things (IoT) devices has further expanded the attack surface, creating new vulnerabilities that malicious actors can exploit. Effective detection of network attacks is crucial for maintaining security and preventing potential damages such as data breaches, financial losses, and disruption of essential services (Khraisat, et al. 2019).

The detection of network attacks can be formulated as a classification problem, where the objective is to distinguish between benign and malicious traffic. The evolution of attack detection methods has progressed from simple rule-based systems to sophisticated machine learning algorithms. Early approaches relied on statistical methods and basic classifiers such as Support Vector Machine (Shon and Moon, 2007). Subsequently, ensemble methods like Random Forest and boosting algorithms demonstrated improved performance (Zhang, et al. 2008). With the advent of deep learning, neural network architectures including Convolutional Neural Networks (CNNs) and Recurrent Neural Networks (RNNs) (Wang, et al. 2017, Vinayakumar, et al., 2017)

Network attack detection presents unique challenges that warrant specific methodological considerations. First, real-world network traffic datasets typically exhibit significant class imbalance, with benign traffic vastly outnumbering malicious instances. This imbalance can bias classifiers toward the majority class (Abdelkhalek & Mashaly, 2023), reducing detection effectiveness for actual attacks. Second, the high-dimensional and heterogeneous nature of network traffic features necessitates models capable of capturing complex relationships without overfitting. As an alternative to traditional deep neural networks (Zhou and Feng, 2017) Deep Forest, which represents an innovative ensemble learning approach. This ensemble learning approach combines multiple decision tree-based models in a cascade structure to achieve deep learning without relying on backpropagation. The development of Deep Forest was motivated by the need for models that could achieve high accuracy while maintaining computational efficiency. Deep Forest offers advantages in addressing these challenges compared to other approaches. Unlike deep neural networks that require extensive parameter tuning and substantial training data, Deep Forest demonstrates greater robustness to overfitting and performs well even with limited samples. Additionally, its ensemble structure provides natural resistance to noise in the input features. Our research investigates the application of Deep Forest to address class imbalance in network attack detection. The results demonstrate that appropriate imbalance treatment combined with Deep Forest significantly improves detection performance across multiple evaluation metrics, outperforming traditional machine learning approaches and even some deep learning models in terms of both accuracy and computational efficiency (Zhang and Zhang, 2023).

The remainder of this paper is organized as follows: Section 2 reviews deep forest and evaluation metrics. Section 3 describes the methodology, including the dataset characteristics, preprocessing steps, feature selection, and implementation details of the models, and presents the experimental results with comparative analysis of different approaches. Finally, Section 4 concludes the paper with a summary of contributions and future extension.

## II. DEEP FORST AND EVALUATION CRITERIA

### A. Deep Forest

The Deep Forest, also known as gcForest was introduced by (Zhou and Feng, 2017) as an alternative to deep neural networks that uses decision tree ensembles instead of neural networks. It aims to achieve performance comparable to deep neural networks while using non-differentiable modules and requiring less hyper-parameter tuning. The underlying logic of gcForest revolves around two main components: the cascade forest structure for deep representation learning and multi-grained scanning for enhanced feature representation.

**Cascade Forest structure**. The cascade forest is the core of the gcForest architecture, designed to perform deep representation learning through a layer-by-layer processing of features. Each layer consists of multiple decision tree ensembles and Each layer receives input features, outputs class probability vectors from each forest, and concatenates these with the original features to form an enriched input for the next layer. Figure 1 shows the structure of Cascade Forest structure.

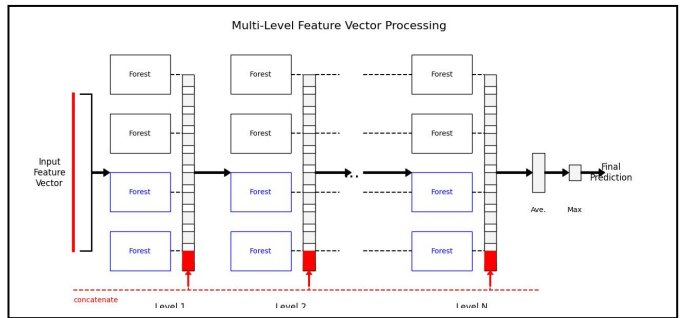

Figure 1. Cascade Forest Structure

**Multi-Grained Scanning**. Multi-Grained Scanning captures local and spatial features by applying sliding windows over the input features, similar to convolution in CNNs. For each window, a set of decision tree forests is trained to generate class probability outputs. These outputs are then concatenated to form a richer feature representation. This process allows the model to detect patterns at different granularities and improves its ability to model complex structures in the data. In 2017, Zhou and Feng describe the process for Multi-Grained Scanning see Figure 2:

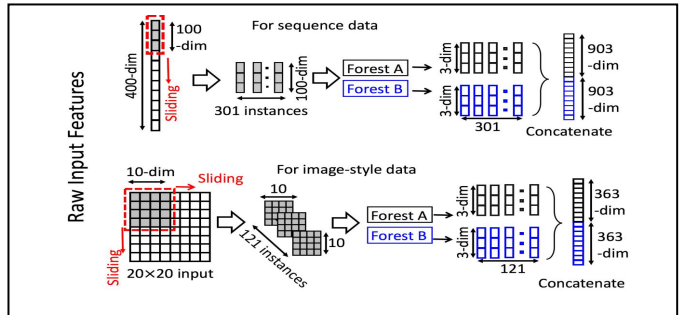

Figure 2. Multi-Grained Scanning

### B. Evaluation criteria

This paper will use five indicators: accuracy, F1 value, precision rate, recall rate and ROC AUC to evaluate the performance of the models. Taking the binary classification problem as an example, the calculation methods of the evaluation criteria are defined as follows:

Accuracy = (TP+TN)/(TP+TN+FP+FN) ,

Precision = TP/(TP+FP),

Recall = TP/(TP+FN),

F1 Score = 2*(Precision*Recall)/(Precision+Recall),

where TP = True Positives, TN = True Negatives, FP = False Positives, FN = False Negatives.

## III. EMPIRICAL ANALYSIS

### A. Dataset Overview

Our dataset, IoT-23, was provided by the Stratosphere Laboratory, containing labeled benign and malicious IoT network flows. The dataset includes detailed annotations to differentiate between legitimate and harmful activities, aiding malware research and detection efforts (Stratosphere Laboratory, Parmisano, et al. 2020). Each row represents a network flow with various features, including:

- Traffic Metadata: Source/destination IP, ports, protocol, service type.
- Packet Characteristics: Number of packets sent/received, bytes transferred.
- Connection State: Indicators of normal or abnormal connection behaviors.

The dataset provides manual labels derived from network analysis at Stratosphere Labs:

- Benign Flows: Normal network activity without any detected intrusion.
- Malicious Flows: Traffic associated with known malware campaigns

### B. Data Preprocessing

To prepare the dataset for modeling, we first handled missing values by replacing missing categorical entries, such as service types, with "unknown" and filling missing numerical values, such as packet counts, with 0. Feature engineering was performed by converting IP addresses into binary format for numerical processing and extracting timestamp-based features, including year, month, day, and hour, to capture temporal patterns in network traffic. Categorical variables such as protocol, connection_state, history, and service were transformed using Label Encoding to convert them into numerical representations suitable for machine learning algorithms. To ensure consistency and mitigate the impact of outliers, we applied Robust Scaling, which standardizes features while being less sensitive to extreme values.

To address class imbalance in the dataset, we implemented three data balancing strategies to ensure a more equitable distribution between benign and malicious network traffic. Synthetic Minority Over-sampling Technique (SMOTE) was applied to generate synthetic samples for the minority class, ensuring that benign and malicious flows were represented in equal proportions. Hybrid Sampling, which combines SMOTE and undersampling, first oversamples the minority class to 12,000 samples and then reduces the majority class to 12,000 samples, maintaining a balanced dataset while preserving data diversity. Lastly, SMOTEENN (SMOTE + Edited Nearest Neighbors) was employed, where SMOTE is used to oversample the minority class, followed by ENN, which filters out noisy samples to improve model robustness and generalization. These techniques help mitigate bias in model training, enhancing the ability to detect malware effectively in network traffic.

## C. Machine Learning Models

We train and evaluate five models—Logistic Regression (Linear Models), Decision Tree and Deep Forest/GCForest (Tree-Based Models), Multi-Layer Perceptron/MLP (Neural Networks), and Support Vector Machine (Kernel-Based Models)—on each of the three sampling methods, resulting in a comprehensive experimental design consisting of 20 distinct experiments. Table I shows model hyperparameters and description.

TABLE I. MODEL HYPERPARAMETERS AND DESCRIPTION

| Model | Hyperparameters |
| --- | --- |
| Support Vector Machine (SVM) | kernel='rbf'   C=0.001   gamma=0.001   probability=True, |
| Logistic Regression | max_iter=1000   penalty='l2' |
| Decision Tree | max_depth=4, min_samples_split=10, ccp_alpha=0.01 |
| Neural Network (MLP) | hidden_layer_sizes=(100,50),alpha =1.0, max_iter=100 |
| Deep Forest (gcForest) | cascade_layer=8, n_cascadeRF=8, window=2 |

## D. Model Result

The experiments and analyses in this study were conducted on Google Colab, a cloud-based Jupyter notebook environment that provides GPU acceleration and pre-installed machine learning libraries. Google Colab was selected for its scalability, ease of use, and seamless integration with Python-based machine learning frameworks. The implementation of data preprocessing, machine learning model training, and evaluation was carried out using Python, leveraging various scientific computing, machine learning, and data visualization libraries. Table II is the model comparison based on ROC AUC and recall performance.

TABLE II. MODEL RESULTS UNDER VARIOUS SAMPLING METHODS AND DIFFERENT MODELS

| Model | Sampling | Accuracy | F1 Score | Precision | Recall | ROC AUC |
| --- | --- | --- | --- | --- | --- | --- |
| Deep Forest | SMOTEENN | 0.98963 | 0.94127 | 0.88906 | 0.99999 | 0.99983 |
| | Hybrid Sampling | 0.98963 | 0.94127 | 0.88906 | 0.99999 | 0.99982 |
| | SMOTE | 0.98963 | 0.94127 | 0.88906 | 0.99999 | 0.99980 |
| | Original | 0.99179 | 0.95238 | 0.91935 | 0.98787 | 0.99942 |
| Neural Network | SMOTE | 0.98963 | 0.93867 | 0.92295 | 0.95494 | 0.99803 |
| | Hybrid Sampling | 0.98934 | 0.93707 | 0.91987 | 0.95494 | 0.99788 |
| | SMOTEENN | 0.98949 | 0.93787 | 0.92140 | 0.95494 | 0.99751 |
| | Original | 0.97595 | 0.83482 | 0.97235 | 0.73137 | 0.98581 |
| Decision Tree | SMOTE | 0.98877 | 0.93596 | 0.88924 | 0.98787 | 0.99361 |
| | Hybrid Sampling | 0.98877 | 0.93596 | 0.88924 | 0.98787 | 0.99361 |
| | SMOTEENN | 0.98877 | 0.93596 | 0.88924 | 0.98787 | 0.99361 |
| | Original | 0.99510 | 0.96964 | 1.00000 | 0.94107 | 0.97054 |
| Logistic Regression | Hybrid Sampling | 0.98675 | 0.92028 | 0.92028 | 0.92028 | 0.99234 |
| | SMOTE | 0.98690 | 0.92135 | 0.91897 | 0.92374 | 0.99218 |
| | SMOTEENN | 0.98690 | 0.92135 | 0.91897 | 0.92374 | 0.99177 |
| | Original | 0.93764 | 0.40440 | 0.98000 | 0.25477 | 0.98709 |
| SVM | Hybrid Sampling | 0.98401 | 0.90220 | 0.91756 | 0.88735 | 0.99394 |
| | SMOTEENN | 0.98200 | 0.89121 | 0.89510 | 0.88735 | 0.99140 |
| | SMOTE | 0.98200 | 0.89121 | 0.89510 | 0.88735 | 0.99140 |
| | Original | 0.95233 | 0.59781 | 1.00000 | 0.42634 | 0.99133 |

Deep Forest demonstrates superior robustness across experimental configurations, achieving perfect recall (0.9999) with SMOTEENN sampling while maintaining the highest ROC AUC (0.99983), indicating exceptional discriminative capability even in the presence of class distribution skewness. When evaluated on the unmodified dataset, it attains an F1 score of 0.95238, exhibiting significantly greater stability than traditional decision tree approaches, which are susceptible to variance-related degradation. Most notably, Deep Forest exhibits remarkable performance consistency across multiple sampling methodologies (SMOTEENN, Hybrid Sampling, and SMOTE), suggesting inherent resilience to sampling strategy selection. This algorithmic stability represents a substantial advantage in operational deployment scenarios characterized by class imbalance, a persistent challenge in network security applications. Deep Forest's capacity to maintain high performance metrics regardless of class distribution interventions demonstrates its intrinsic ability to effectively model the underlying data structure while resisting overfitting, properties particularly valuable in dynamic threat landscapes where attack signatures continuously evolve.

## E. Feature Importance

The proposed feature selection methodology implements a rank-based ensemble approach that integrates XGBoost and Random Forest algorithms. Both models independently rank features according to their importance metrics, which are then consolidated into a unified framework. Features absent from either method's importance list receive penalized rank values. The final feature significance is determined by calculating the mean rank position across both algorithms, with lower values indicating higher importance. This approach effectively leverages the complementary strengths of both algorithms while normalizing their outputs through rank transformation rather than raw importance scores, thereby eliminating scale discrepancies and producing a more robust feature subset for network attack detection. Table III is our final 10 feature based on the importance:

TABLE III. FEATURE IMPORTANCE WITH DIFFERENT MODEL AND FINAL RANK

| Feature | Method 1: XGBoost | Method 2: Random Forest | Final Rank[a] |
| --- | --- | --- | --- |
| protocol | 0.6095 (Rank1) | 0.3464 (Rank1) | Rank 1 |
| origin_port | 0.0321 (Rank4) | 0.2335 (Rank2) | Rank 2 |
| connection_state | 0.2604 (Rank2) | 0.073 (Rank4) | Rank 3 |

| Feature | Method 1: XGBoost | Method 2: Random Forest | Final Rank[a] |
|---|---|---|---|
| response_host_binary | 0.0505 (Rank3) | 0.0695 (Rank5) | Rank 4 |
| response_port | 0.0149 (Rank6) | 0.077 (Rank3) | Rank 5 |
| original_ip_bytes | 0.0186 (Rank5) | 0.0628 (Rank6) | Rank 6 |
| hour | 0.0042 (Rank7) | 0.0378 (Rank8) | Rank 7 |
| service | 0.0032 (Rank8) | 0.0127 (Rank10) | Rank 8 |
| history | 0.0008 (Rank13) | 0.0464 (Rank7) | Rank 9 |

a. Rank is determined by calculating the mean rank position across both algorithms, with lower values indicating higher importance.

## IV. CONCLUSION AND FUTURE EXTENSION

In conclusion, our comprehensive evaluation of machine learning approaches for network attack detection demonstrates the superior efficacy of Deep Forest across multiple performance dimensions. The model exhibits exceptional robustness to dataset imbalance, maintains consistent performance across diverse sampling methodologies, and achieves optimal classification metrics in real-world detection scenarios. Unlike alternatives that excel in isolated metrics but falter in others, Deep Forest provides a balanced performance profile with high recall, precision, and AUC values—critical for operational security deployments where false negatives could lead to compromised systems while false positives create unsustainable alert volumes. Its computational efficiency and minimal hyperparameter requirements further enhance practical utility. These findings suggest Deep Forest represents a significant advancement for network security applications, offering a reliable detection framework capable of addressing the complex challenges presented by modern cyber threats in dynamic network environments.

The promising results of Deep Forest in network attack detection open several avenues for future research and application. Integration with explainable AI techniques could enhance interpretability while maintaining high performance, addressing a critical need in security applications where understanding decision rationales is essential.